%% file: main.tex
\pdfoutput=1
\documentclass[11pt]{article}

\usepackage{setspace} %

\usepackage[]{acl}

\usepackage{times}
\usepackage{latexsym}
\usepackage{amsmath,amssymb}
\usepackage{dsfont}
\usepackage{bm} %
\usepackage{xfrac}
\usepackage{soul}
\usepackage[T1]{fontenc}
\usepackage[utf8]{inputenc}
\usepackage{microtype}
\usepackage{inconsolata}
\usepackage{pifont}

\usepackage{tabularx}
\usepackage{booktabs,tabularx}
\usepackage[textwidth=2.25cm]{todonotes}
\usepackage[capitalise,noabbrev]{cleveref}
\usepackage{xspace}
\usepackage{enumitem}

\usepackage{makecell}

\usepackage{marginnote}

\usepackage{xcolor} %
\usepackage{soul}   %

\newcommand{\toshipone}{{ToShip21}\xspace}
\newcommand{\toshiptwo}{{ToShip23}\xspace}

\def\parcite#1{\citep{#1}} %
\def\perscite#1{\citet{#1}} %
\def\inparcite#1{\citealp{#1}} %

\definecolor{TodoColor}{rgb}{1,0.7,0.6}

\definecolor{RedA}{rgb}{0.7,0.3,0.3}
\definecolor{BlueA}{rgb}{0.3,0.3,0.7}
\definecolor{RoyalBlue3}{RGB}{58,95,205}
\definecolor{Green4}{RGB}{0,139,0} 

\let\svthefootnote\thefootnote
\newcommand\blankfootnote[1]{%
  \let\thefootnote\relax\footnotetext{#1}%
  \let\thefootnote\svthefootnote%
}

\newcommand{\hlc}[2][yellow]{{%
    \colorlet{foo}{#1}%
    \sethlcolor{foo}\hl{#2}}%
}

\newcommand\comet[2][]{Comet$_{#2}^\textrm{#1}$\xspace}
\newcommand\cometkiwi{CometKiwi$_{22}^\textrm{QE}$\xspace}
\newcommand\bleu{{BLEU}\xspace}
\newcommand\chrf{{ChrF}\xspace}
\newcommand\spbleu{spBLEU$^\textrm{200}$}
\newcommand\bleurt{{BLEURT$_{20}$}\xspace}
\newcommand\bleurtdef{BLEURT$_\text{default}$\xspace} %
\newcommand\xcometxxl{{xCOMET$_\textrm{XXL}$}\xspace}

\title{
Navigating the Metrics Maze: \\
Reconciling Score Magnitudes and Accuracies
}

\newcommand{\microsoftsigil}{$^1$}
\newcommand{\ethsigil}{$^2$}

\author{Tom Kocmi\microsoftsigil \qquad Vilém Zouhar\ethsigil \qquad Christian Federmann\microsoftsigil \qquad Matt Post\microsoftsigil \\[1em]
  \hspace{2.1cm} \microsoftsigil{}Microsoft \hspace{5cm} \ethsigil{}ETH Zürich\\
  \texttt{\{tomkocmi,chrife,mattpost\}@microsoft.com}
  \qquad \texttt{vzouhar@ethz.ch} 
  }

\begin{document}

\maketitle

\begin{abstract}

Ten years ago, a single metric, BLEU, governed progress in machine translation research.
For better or worse, there is no such consensus today, and consequently it is difficult for researchers to develop and retain intuitions about metric deltas that drove earlier research and deployment decisions.
This paper investigates the ``dynamic range'' of a number of modern metrics in an effort to provide a collective understanding of the meaning of differences in scores both within and among metrics; in other words, we ask \emph{what point difference $x$ in metric $y$ is required between two systems for humans to notice?}
We conduct our evaluation on a new large dataset, \toshiptwo, using it to discover deltas at which metrics achieve system-level differences that are meaningful to humans, which we measure by pairwise system accuracy.
We additionally show that this method of establishing delta-accuracy is more stable than the standard use of statistical p-values in regards to testset size.
Where data size permits, we also explore the effect of metric deltas and accuracy across finer-grained features such as translation direction, domain, and system closeness.

\end{abstract}

\footnotetext[0]{
Paper code:\hspace{5.3mm}
\href{https://github.com/kocmitom/MT-Thresholds}{github.com/kocmitom/MT-Thresholds}\hspace{2mm}\null \\
\null\hspace{5.6mm}Online tool: \hfill
\href{https://kocmitom.github.io/MT-Thresholds/}{kocmitom.github.io/MT-Thresholds}\hspace{2mm}
}

\section{Introduction}

A decade ago, the BLEU metric served as the default metric for machine translation evaluation.
It was not without its criticisms \parcite{hovy-ravichandran-2003-holy, callison-burch-etal-2006-evaluating, belz-reiter-2006-comparing} or compelling alternatives \parcite{banerjee-lavie-2005-meteor, popovic-2015-chrf}, but a combination of adequate performance, robustness to new languages, simplicity, understandability---and also inertia---helped it retain this position.
This is no longer the case.
BLEU's deficiencies quickly became apparent as deep learning approaches to machine translation replaced the earlier symbolic paradigms \parcite{mathur-etal-2020-tangled}.
Today, a number of metrics---themselves deep-learning based---compete in an ecosystem where there is no longer any dominant, default metric.

\begin{figure}[t]
    \centering
    \includegraphics[width=\linewidth]{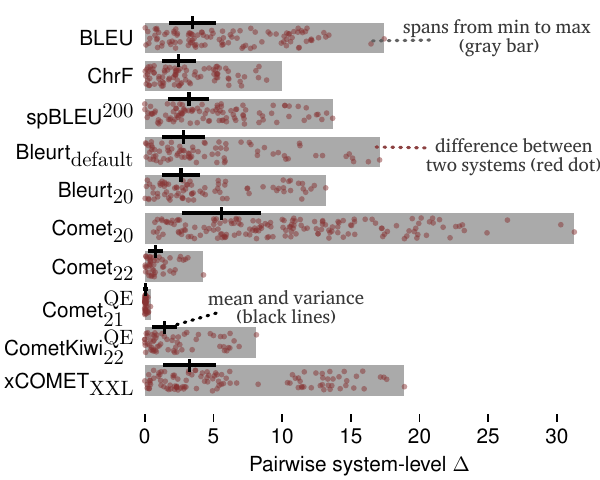}
    \vspace*{-6mm}
    \caption{Distribution of pairwise system deltas for each metric over all systems from WMT22. Gray rectangles show min-max range which is vastly different between metrics. Standard deviations (black lines) also differ.
    }
    \label{fig:score_distribution}
    \vspace{-4mm}
\end{figure}

This situation creates a problem for researchers working to keep abreast of developments in the field.
Different metrics, including different models within the same metric family, have different \emph{dynamic ranges}, i.e., the range of scores one can expect to see.
Furthermore, the \emph{metric delta}, i.e., the score difference signifying a meaningful change in performance between two systems, also varies across metrics.
It is perhaps understandable that some practitioners therefore continue to use BLEU, as well, if only to ground their understanding.

This paper attempts to introduce some order and clarity into this situation.
We make use of a large, new human evaluation dataset, \toshiptwo, to compare the score ranges of metrics on a large number of systems against pairwise system-level accuracy.
Importantly, we break down these accuracy scores into bins based on metric deltas, which allows us to determine accuracies for each metric as a function of the score differences between two systems.
This provides a measure of confidence in the output that is stable across testset size, in contrast to standard statistical significant testing, which becomes more stable as testset size grows.
We release a tool that allows a user to easily compare accuracies at different threshold across metrics.\footnotemark[0]

In this work we:
\begin{enumerate}[noitemsep,topsep=0mm,left=4mm,label={\arabic*)}]
\item[\S\ref{sec:delta_size}] Empirically investigate the estimated accuracy for multiple metrics, human ability to perceive quality difference;
\item[\S\ref{sec:aligning_metrics}] Provide thresholds for popular metrics to help reviewers and practitioners interpret results;
\item[\S\ref{sec:domain_dataset}] Validate our estimated accuracies on WMT testsets and \S\ref{sec:language_pairs} investigate the effect of different language groups;
\item[\S\ref{sec:iterated_systems}] Show that string-based metrics, such as BLEU, should never be used to evaluate unrelated systems;
\item[\S\ref{sec:testset_size}] Show that statistical significance testing is insufficient to determine model improvement especially as it is affected by the testset size, but is important for small deltas;
\item[\S\ref{sec:full_metric_evaluation}] Assess quality of automatic metrics over 6530 system pairs;
\item[\S\ref{sec:recommendations}] Summarize recommendations for machine translation evaluation.
\end{enumerate}

\section{Experimental Setup}
\label{sec:experimental_setup}

\paragraph{Data.}
We perform experiments related to evaluation of MT outputs based on a proprietary dataset \emph{ToShip23} which is of a magnitude larger than any publicly available data and enables more fine grained glimpse into the metrics behaviour.
The dataset is an extended version of \toshipone dataset \citep{kocmi-etal-2021-ship} with details described in \Cref{app:ToShip2}.
We also use data from the annual WMT evaluation campaigns to validate our results, specifically the metrics shared task \citep{freitag-etal-2022-results, freitag-etal-2023-results}, to make results replicable.
We only use MQM \parcite{freitag-etal-2021-experts} and DA+SQM \parcite{kocmi-etal-2022-findings} subset of human evaluated systems because reference-based DA \parcite{bojar-etal-2016-findings} is suboptimal for the evaluation of modern MT systems \citep{freitag-etal-2022-results}.
See \Cref{tab:data_overview} for an overview of dataset sizes.

\begin{table}[htbp]
\addtolength{\tabcolsep}{-0.25em}
\resizebox{\linewidth}{!}{
\input{generated/dataset_overview.tex}
}
\vspace{-1mm}
\caption{Sizes and coverage for the human annotated datasets used in this work.}
\label{tab:data_overview}
\vspace{-4mm}
\end{table}

\paragraph{Investigated Metrics.}
We evaluate the most frequently used metrics in machine translation: BLEU \citep{papineni-etal-2002-bleu}, ChrF \citep{popovic-2015-chrf}, spBLEU \parcite{goyal-etal-2022-flores}, BLEURT \citep{sellam-etal-2020-bleurt}, COMET \citep{rei-etal-2020-comet}.
BLEU and ChrF are n-gram matching heuristics while the rest uses a parametric model to produce a segment-level score of a translation. 
\comet[QE]{21} and \cometkiwi{} are special cases which do not require a reference.
We do not include any LLM-based metrics \citep{fernandes-etal-2023-devil, kocmi-federmann-2023-gemba} which are not replicable because of non-publicly available models.
Find the specific models, implementation details, and our selection rationale in \Cref{app:metrics_details}.

\paragraph{Metric Delta.}
We focus solely on the pairwise system ranking: deciding which system is better based on a system-level score (usually average of all segment-level scores) difference between two systems.
We refer to this as \textit{metric delta} ($\Delta$).

\paragraph{Pairwise Accuracy.} 
To test the correlations between automatic metrics and human judgement, we use pairwise accuracy \citep{kocmi-etal-2021-ship}: how many system pairs does the metric rank the same way as humans over the total number of system pairs in the dataset.
Formally:
\begin{align}
\mbox{Acc} = \frac{|\mbox{sign}(\mbox{metric} \Delta) = \mbox{sign}(\mbox{human} \Delta)|}{|\mbox{all system pairs}|} \quad.\nonumber
\end{align}

\section{Unifying Metric Ranges}

We first look at the ``dynamic ranges'' exhibited by different metrics across our datasets.
We ground these deltas in human scores by comparing pairwise system-level accuracy at different thresholds of delta.
With this, we are able to establish a table of average metric deltas for different accuracy levels, and build a simple model that maps any metric into the unified space of estimated accuracies.

\subsection{Various Ranges for Metric Deltas}
\label{section:unified}

\Cref{fig:score_distribution} depicts the distribution of system-level score deltas for various metrics.
Some metrics have similar ranges, such as \chrf{} and \bleu{}, while others use a much larger score range (\comet{20} has ${\sim}5\times$ higher deltas to \bleu{}) or lower score range (\comet[QE]{21} has ${\sim}\sfrac{1}{5}$ range of \bleu{}).

In addition to the wide ranges of scores, we also observe that metrics do not always have the same direction or agreement with human judgment, which results in their different performance as measured via accuracy (see \Cref{app:onebleu} for more details).

It may be tempting to attempt to bring together these score ranges onto a single scale, say by linear interpolation, perhaps towards \bleu{} scale.
But reconciling metrics by projection is not possible, due to an obvious point: metrics differ not just in the range of their scores, but in their accuracies.
To better understand the problem, we look next into what are the implications of different levels of metric deltas.
Specifically, we investigate how different delta correspond to humans being able to differentiate systems.

\subsection{Accuracy of Metric Deltas}
\label{sec:delta_size}

Many factors affect metric behavior:
\begin{itemize}[noitemsep,left=0mm,topsep=0mm]
    \item Each metric weights various phenomena differently, especially fluency and adequacy \parcite{amrhein-etal-2022-aces, karpinska-etal-2022-demetr}.
    \item The reliability of metrics differs when compared to humans \citep{mathur-etal-2020-results,freitag-etal-2021-results,freitag-etal-2022-results,freitag-etal-2023-results, kocmi-etal-2021-ship}.
    \item Reference-based metrics are affected by the quality of human references \citep{freitag-etal-2023-results, zouhar2024quality}.
\end{itemize}

\begin{table*}[htbp]
  \centering
  \setlength\tabcolsep{4mm}
  \resizebox{\linewidth}{!}{
  \input{generated/main_thresholds.tex}
  }
  \caption{Thresholds and estimated accuracies for each metric on \toshiptwo dataset averaged across all language pairs.
  For example, when requiring 90\% of decisions be the same as humans, improvement needs to be $\geq$3.05 \chrf{}, $\geq$0.85 \cometkiwi{}, and \bleu{} never reaches this accuracy threshold.}
  \label{tab:main_thresholds}
  \vspace{-3mm}
\end{table*}

\begin{figure}[t]
  \centering
  \includegraphics[width=\linewidth]{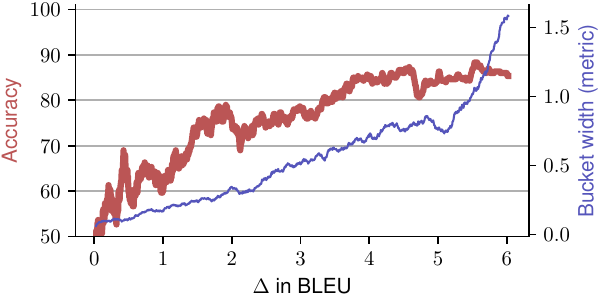} 
  \includegraphics[width=\linewidth]{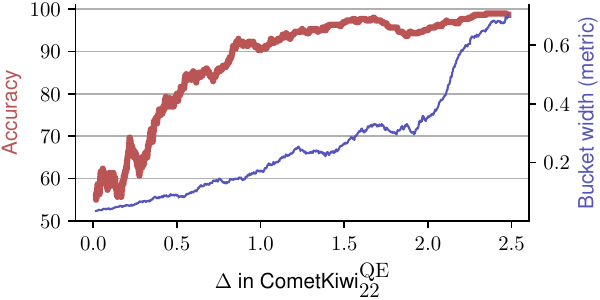} 
  \caption{What pairwise \textcolor{RedA}{accuracy (left-y-axis)} to expect when seeing given certain acceptance threshold (x-axis).
  The \textcolor{BlueA}{bin width (right-y-axis)} shows the width of the bin for metric delta that contains 300 system pairs.
  }
  \label{fig:expected-accuracy}
\end{figure}

\noindent The pairwise accuracy as usually reported \parcite{kocmi-etal-2021-ship, freitag-etal-2023-results} represents a value over the full dataset for all system-pair metric deltas.
It does not take into consideration the \textit{size} of the delta between systems, which heavily affects the accuracy; that is, whether the metric gap between two systems was large or small.
However, this information is important in establishing equivalency of deltas across metrics.

To investigate this, we use a binning approach on the \toshiptwo testset.
Pairwise system deltas are sorted, and for each delta level, we group the closest 300 pairs into a same bin.
For each bin, we plot the mean delta for that bin against the system-level pairwise accuracy.\footnote{\Cref{app:bucket_size} investigates other sizes of bins than selected 300 system pairs.}

\Cref{fig:expected-accuracy} depicts this information for both \bleu and \cometkiwi.
The red line shows that we need around 1.3 \bleu delta to reach 70\% pairwise accuracy and 3.5 \bleu to reach 80\% accuracy against the human judgments.
Because \bleu{} is not a reliable metric, it never reaches 90\% accuracy with humans, even for deltas as high as 6 \bleu{} points.
In contrast, \cometkiwi{} reaches 90\% accuracy already at around 0.9 points and gets close to 100\% accuracy past 2 \cometkiwi{} points.

Our use of fixed-size bins introduces a caveat into the evaluation.
Because our data points do not have a uniform delta distribution, the ``width'' of each bin (defined as the difference between the smallest and largest delta) grows as we move towards larger deltas, where data points are sparser.
This width is depicted by the blue line in \Cref{fig:expected-accuracy}.
As we increase the delta, there are fewer and fewer systems with as large delta and thus we need to take system pairs that are farther from the investigated delta.
For example, for calculating the pairwise accuracy of 1 \bleu point, we take system pairs with a delta of 1 $\pm$0.1 (half of 0.2), while for 3 \bleu{} the width of a bin is 3 $\pm$0.25 points.
The bin width mainly affects the tail of the evaluation.

As our evaluation is empirical, it is heavily affected by the underlying systems and the lines fluctuate.
In the next section, we try to fit a smooth line to abstract the results, followed by discussion which phenomena affect the pairwise accuracy.

\subsection{Aligning Metrics on Accuracy}
\label{sec:aligning_metrics}

Practitioners might be interested in getting an intuition behind a particular metric delta, e.g., +0.10 of \comet{22} and how such delta relates to other metrics that they are familiar with.
Clearly, the higher the delta, the more likely that human raters would also notice the quality difference between systems.
It remains unclear what delta is enough to warrant acceptance.
To this end, we use the estimated accuracy results introduced in previous subsection.
As the estimated accuracy line is noisy, we fit a curve through the data and use it to derive thresholds for comparing various metric deltas.

We use a parametrized sigmoid to fit a curve through the data.
The choice of the sigmoid function is arbitrary and based on visual similarity and the feature that it converges towards fixed point and thus is bounded. This is a desired feature representing that each metric has a different overall reliability.
We parameterize it using two variables $\varphi$ and fit it with damped least square algorithm \citep{levenberg1944method}.
The function is defined as:
\begin{align}
f(x) = \frac{\varphi_1}{1+\textrm{exp}(-\varphi_2\cdot x)} \quad . \nonumber
\end{align}

\begin{figure}[t]
  \centering
  \includegraphics[width=0.49\linewidth]{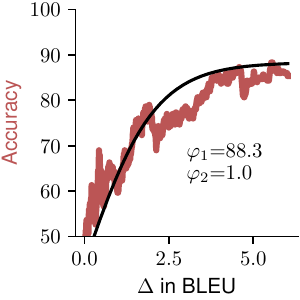} 
  \hfill
  \includegraphics[width=0.49\linewidth]{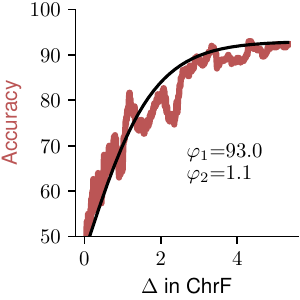} 
  \\
  \raisebox{-1mm}{\includegraphics[width=0.49\linewidth]{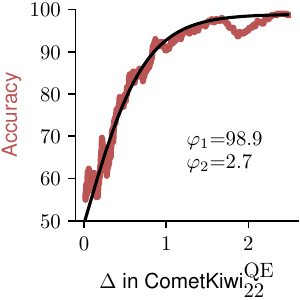}}
  \hfill
  \includegraphics[width=0.49\linewidth]{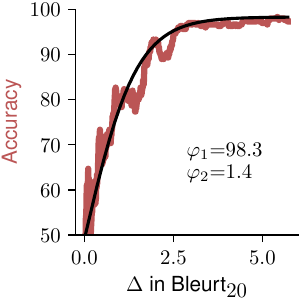} 
  \caption{
  Empirical pairwise \textcolor{RedA}{accuracies} for various metrics with a fitted sigmoid curves on \toshiptwo dataset. All metrics are in \Cref{fig:expected-accuracy-with_sigmoid_big}.
  }
  \label{fig:expected-accuracy-with_sigmoid}
  \vspace{-3mm}
\end{figure}

The resulting fit is visualized in \Cref{fig:expected-accuracy-with_sigmoid}.
Although not perfect, it offers insight into the metric delta behaviour, specifically comparing different different deltas' estimated accuracy.
We use the sigmoid functions to calculate estimated accuracy for various levels of delta in \Cref{tab:main_thresholds}.
This is the core result of our work and helps in understanding how different metrics compare to each other.

For example, an improvement of 1.06 \bleu{} has the same estimated accuracy (65\%) as the 0.24 \cometkiwi{}, while 3.35 \bleu has the same estimated accuracy as 0.67 \cometkiwi{}. 
And +1 improvement on \cometkiwi signals that in >90\% scenarios, human annotators would agree with the ranking of \cometkiwi{}, while BLEU never reaches this level of agreement.
Note that estimated accuracies are empirical from a given \toshiptwo{} dataset.
Therefore, we do not claim that +0.56 \comet{22} yields 80\% accuracy for all scenarios but rather that it is as accurate as +2.34 \bleu{}.

\begin{figure}[t]
  \centering
  \includegraphics[width=\linewidth]{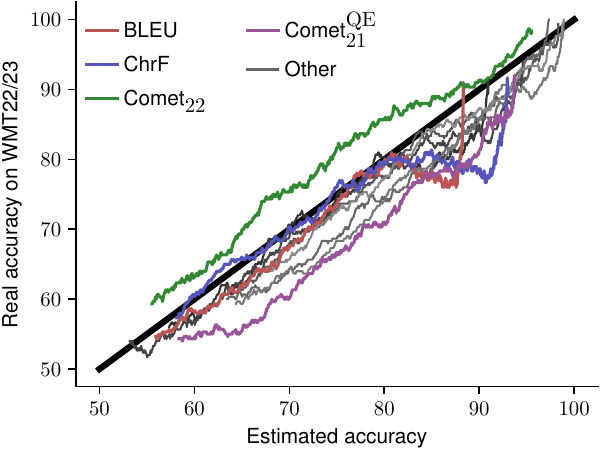}
  \caption{
  Testing the validity of thresholds devised on \toshiptwo with WMT datasets. In a scenario without noisy data, we would expect the real accuracies to match the estimated accuracies (the black line).
  See detailed per-metric breakdown in \Cref{fig:test_on_wmt_big}.
  }
  \label{fig:test_on_wmt}
\end{figure}

As these thresholds are combined for all scenarios, we dive in the next section into validating out results on public WMT dataset, followed with investigation of what affects the metric delta and how reliable the comparison is in different settings.

\section{Factors Affecting Metric Deltas}

We have \textit{empirically} derived the estimated accuracy for various metrics.
In this section, we investigate factors that affect metric delta and show how reliable the thresholds remain under these factors.
These include the testset size, dataset and domain selection, and translation direction.

Additional factors could influence the metric delta, but we lack the data to evaluate these aspects. A key consideration is whether the metric delta is contingent on the underlying absolute values. In other words, we need to determine if a +1 \bleu{} delta varies in reliability based on these absolute values. For instance, does the impact of moving from 20 to 21 \bleu{} differ significantly from a shift from 60 to 61 \bleu{} in different system pairs?

\begin{figure}[t]
  \centering
  \includegraphics[width=0.97\linewidth]{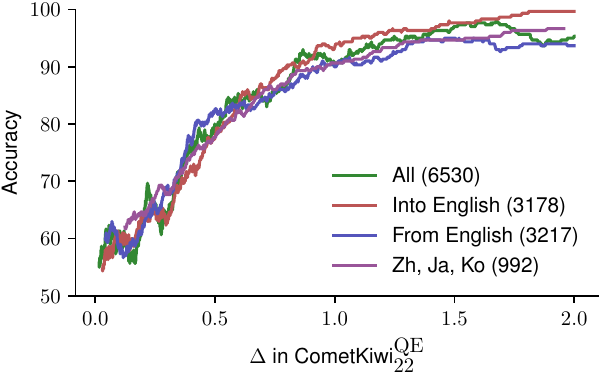}
  \includegraphics[width=0.97\linewidth]{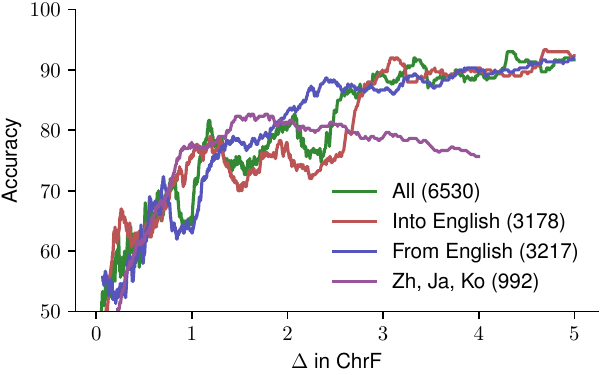}
  \caption{
  Comparison of pairwise accuracy on \toshiptwo dataset when comparing into English, out-of-English, and Chinese, Japanese, Korean language pairs separately. The count shows total number of system-pairs in the evaluation.
  See other metrics in \Cref{fig:XE_EX_comparison_big}.
  }
  \label{fig:XE_EX_comparison}
\end{figure}

\subsection{Different Domains and Datasets}
\label{sec:domain_dataset}

We derived the thresholds from \toshiptwo. Now, we validate them on WMT data to show how well they transfer.
To address the relatively small size of WMT, we first combine the WMT 2022 and 2023 datasets, yielding 1414 system pairs.
This dataset contains different set of segment sources and domains, and was evaluated with mix of MQM and DA+SQM human evaluation protocols.
In order to test the thresholds, we take scores for all WMT system pairs and convert them into estimated accuracies via devised thresholds. 
For each estimated accuracy level, we take the closest 300 system pairs and calculate the real accuracy on WMT data.
If the mapping would be perfect and we had enough samples, the estimated accuracy would match the real accuracy for each investigated level.

We show the results in \Cref{fig:test_on_wmt}.
In the ideal case, we would expect the real accuracies and estimated accuracies to match; however, the noise from empirical data affects the results.
Some metrics are consistently underestimated, such as \comet{22}, which has higher real accuracies on WMT dataset that the estimated accuracies.
On the other hand, \comet[QE]{21} has much lower accuracies on WMT data and our thresholds overestimate it. 

Overall, the trend is clear and the thresholds normalize all metrics into a shared space of estimated accuracies.
Therefore, we advise reporting accuracy when presenting results, together with significance testing and metric delta.

\subsection{Language Pair}
\label{sec:language_pairs}

Notoriously, a large gap in absolute \bleu scores exists between languages \citep{denoual-lepage-2005-bleu,post-2018-call}.
This reflects properties like data sizes, attention progress in different languages, and target-side morphological complexity.

Unfortunately, there is not enough data to examine each language pair individually.
Instead, we bin languages into two groups, \emph{into-English} (XE) and \emph{out-of-English} (EX) language pairs, which does leave us with enough data in the \toshiptwo dataset.
In addition, we separate system pairs containing Chinese, Japanese, or Korean (CJK) together.

\Cref{fig:XE_EX_comparison} show the accuracy with a subset of system pairs depending on a languages.
There is some fluctuation between XE and EX, but the behaviour is comparable.
This is interesting, since most of the underlying testsets have authentic source (e.g., not using testset in reverse direction, \inparcite{toral-etal-2018-attaining}).
The CJK group does also perform similarly for \cometkiwi, but not for \chrf. 
This shows the thresholds are invalid for all metrics and scenarios and are affected by whether metrics evaluate all language similarly or not.

\subsection{Iterated versus Unrelated Systems}
\label{sec:iterated_systems}

Another main difference that affects the evaluation is if the systems are closely related.
Key point of distinction is between \textit{iterated systems} (a baseline system against specific improvements, produced by the same research group) or \textit{unrelated system} (for example, WMT yearly evaluation which comes from different teams and systems produce vastly different translations).
It has long been known that surface metrics like BLEU work best when evaluating closely-related iterated systems \citep{callison-burch-etal-2006-evaluating}.
It may be easier for both metrics and humans to distinguish an iterated system over its baseline, because comparing unrelated systems adds a difficulty of weighting different styles and errors.

\begin{figure}[t]
  \centering
  \includegraphics[width=0.97\linewidth]{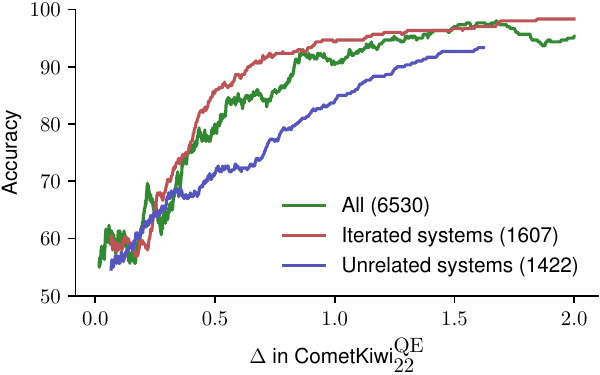}
  \includegraphics[width=0.97\linewidth]{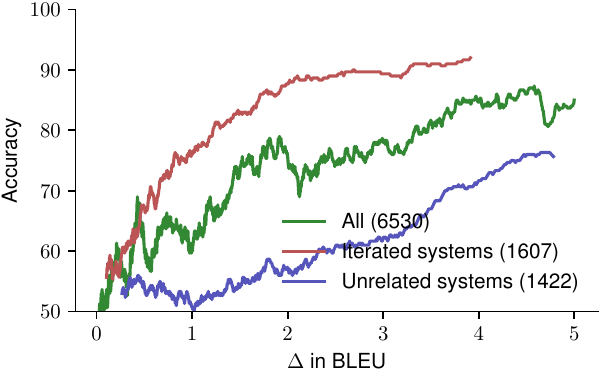}
  \caption{
  Comparison between iterated and unrelated systems on \toshiptwo.
  See other metrics in \Cref{fig:unrelated_systems_big}.
  }
  \label{fig:unrelated_systems}
\end{figure}

To investigate this, we use the system labels of \toshiptwo dataset, where some system pairs are baseline model and it's improved iterated model, while other system pairs are completely unrelated and developed by different teams, similarly to WMT evaluation.
\Cref{fig:unrelated_systems} confirms the assumption that unrelated systems are much harder to evaluate and that the metric behaves differently. Therefore, automatic metrics are better to rank iterated systems than unrelated systems. While pretrained metrics, such as \cometkiwi{}, seems to be robust enough for comparing both types of system pairs, other metrics such as \bleu have much harder time to distinguish unrelated systems. This effect should be investigated to larger detail in future work.

For example, +2 BLEU on iterated models has an accuracy with humans of about 90\%, the same +2 BLEU on unrelated systems are barely better than toss of a coin ($\approx$55\%). This shows, that some metrics (specifically BLEU, ChrF, spBLEU) should not be used to evaluate unrelated systems. 
This findings was also suggested by \perscite{berg-kirkpatrick-etal-2012-empirical}, who showed that you need to get about one third larger BLEU improvements for unrelated systems to reach the same p-value.

Therefore, string-based metrics, such as BLEU, ChrF, or spBLEU, should never be used to compare unrelated systems.

\subsection{Testset Size}
\label{sec:testset_size}

\begin{figure*}[htbp]
  \centering
  \includegraphics[height=4.11cm,trim={0 0 0mm 0},clip]{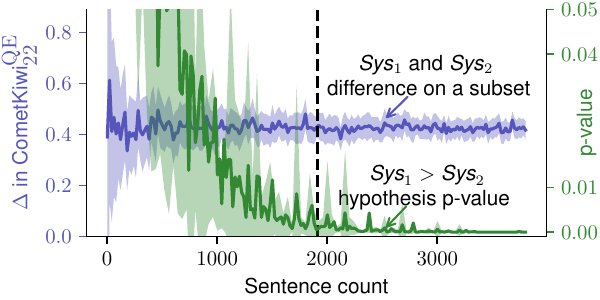}
  \hfill
  \includegraphics[height=4.11cm,trim={7mm 0 0 0},clip]{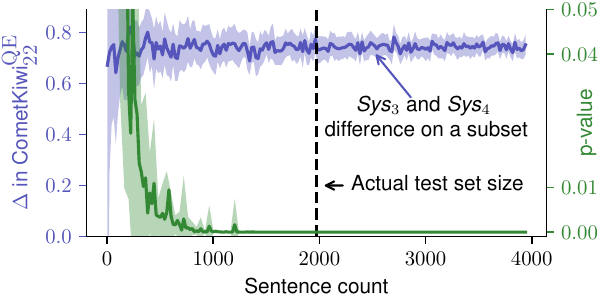}
  \\
  \includegraphics[width=\linewidth]{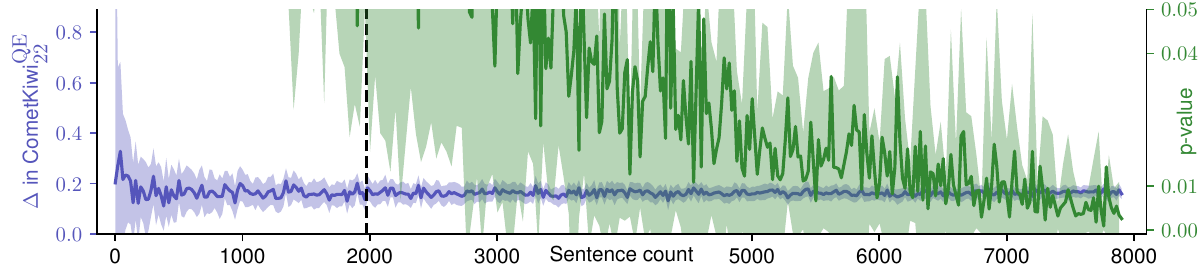}
  \caption{Three system pairs on different languages from WMT23 scored by \cometkiwi. The \textcolor{RoyalBlue3}{blue line} is the average system delta for given testset size and \textcolor{Green4}{green line} is the associated p-value.
  Values to the right of the dashed line are supersampled and shaded areas are 99.9\% confidence intervals from t-distribution.
  The metric delta does not change much while the p-value goes down with higher subset size.
  }
  \label{fig:testset_size}
\end{figure*}

Another phenomena that may affect the system delta is the number of sentences in the parallel testset used to evaluate pair of systems.
Common wisdom says that the testset should be as large as possible.
We ask if increasing testset size affects the system delta and its statistical significance.

To examine how testset size affects the metric delta, we take a system pair and sample testsets with increasing number of sentences.
For each sample, we calculate \cometkiwi delta and p-value using paired Student's t-test \parcite{mathur-etal-2020-tangled}.
We sample with repetition various testset sizes.
For each testset size, we plot the average metric delta (or p-value respectively) over 50 runs together with the confidence interval.

From \Cref{fig:testset_size}, the metric delta fluctuates but keeps being mostly constant.
The variance of the metric delta is higher for small testset sizes (under 500 segments). 
On the other hand, the p-value associated to the comparison hypothesis goes down simply by having a larger testset, phenomena shown for MT by \perscite{berg-kirkpatrick-etal-2012-empirical}.

This is a natural phenomenon of statistical significance testing \citep{greenland2016statistical}.
P-values decrease with an increasing sample size, assuming the null hypothesis does not hold.
This is due to the increase in statistical power---the probability that the test correctly rejects the null hypothesis when it is false.
Should the null hypothesis hold perfectly, which is rarely the case, increasing the sample size would not systematically affect the p-values.
Therefore, it is possible to claim a statistically significant improvement over a baseline model even with a small metric delta, which might not be noticeable by humans, just by using a large-enough testset.
This conclusion is not an argument against the use of statistical significance testing, which remains important, especially when observing smaller deltas.

Overall, this shows that metric delta is stable under different testset sizes, while statistical significance testing is affected by it.
We assumed to be adding sentences from the same distribution.
The metric delta can be manipulated by adding segments that are more difficult than the rest.

\section{Discussion}

\subsection{Best-performing Metrics}
\label{sec:full_metric_evaluation}

With the \toshiptwo dataset, we can also calculate total pairwise accuracy over all system pairs to devise which metrics perform the best on the (to date) largest dataset of MT human evaluation. 
We follow the same evaluation as in Table 2 from \perscite{kocmi-etal-2021-ship}.
Twice as large dataset than \toshipone, extended by state-of-the-art systems from 2022 and 2023, we can see how metrics perform on system-level rankings.
\Cref{tab:accuracy_results} shows that the best performing metric over the \toshiptwo dataset is \cometkiwi{} by a small margin over \xcometxxl{}. \cometkiwi{} is a quality estimation metric, which has an additional bonus of not being affected by reference bias.

\begin{table}[htbp]
  \centering
  \resizebox{\linewidth}{!}{
  \input{generated/accuracy_results}
  }
  \caption{A pairwise accuracy over all system pairs from \toshiptwo{} and two subsets depending on the year of evaluation. The results of MQM subset of WMT23 \citep{freitag-etal-2023-results}.}
  \label{tab:accuracy_results}
\end{table}

\noindent Additionally, we notice the overall accuracy dropped for all metrics in the last two years.
This does not necessarily signify a drop in metric performance, but may have several other explanations:
\begin{itemize}[left=0mm,topsep=0mm,noitemsep]
    \item \textbf{Different systems}: Newer architectures or systems are closer to each other in performance, thus harder to evaluate by humans
    \item \textbf{New testsets}: While the 2019-2021 contains only two domains, the newer data have been evaluated on a much larger set of domains, where some domains may be challenging for metrics
    \item \textbf{Human bias}: The evaluation protocol changed, which may have shifted annotator's scoring patterns.
\end{itemize}
\smallskip

\noindent
However, the absolute pairwise accuracy is less important than the ranking of metrics, as it is heavily affected by the system pairs. We compare to MQM subset of \citet{freitag-etal-2023-results}, which ranks metrics in similar order supporting our findings. 
There are some notable differences, such as \comet[QE]{21} ranking as the worst metric in WMT, while \bleurtdef{} is the worst in \toshiptwo{}. 
Since many aspects of the evaluation are different, we do not dive into a comparison, but rather highlight the overall picture.
\toshiptwo corroborates that QE metrics have reached the quality of reference-based metrics, as well as the (already well-established) fact that lexical-based metrics are not useful for evaluating high-resource MT models these days.

\subsection{Recommendations for MT Evaluation}
\label{sec:recommendations}

We conclude with a list of recommendations for automatic MT evaluation:

\begin{itemize}[left=0mm,noitemsep,topsep=1mm]
    \item Use \cometkiwi as the main metric.
    In addition to its better performance, as a quality estimation metric, it is not affected by references.
    \item Use at least one additional metric of a different type; e. g. \bleurt{}, which is reference-based and uses a different architecture from Comet.
    \item For each metric delta, report estimated accuracy to help align reliability of used metrics.
    \item Do not use BLEU, ChrF, or spBLEU to evaluate unrelated systems.

\end{itemize}

In addition, employ caution when using the same metric for evaluation that was used during training, as this practice may lead to artificially inflated results.
For instance, it is advisable not to evaluate with the same metric used for Minimum Bayes Risk Decoding \parcite{freitag-etal-2022-high}, QE metric used for corpus filtering \parcite{peter-etal-2023-theres}, or avoid using metrics built on the same model as the translation system because LLMs tend to favor outputs generated by themselves \parcite{liu2023llms}.

\section{Related Work}

The closest work to ours is \citet{lo2023beyond}, who investigate the relationship between metric deltas and the p-value of human ranking, concluding that not even 2 \bleu points reliably correspondent to human judgement.
This aligns with our work that two \bleu points reaches an estimated accuracy of only 77.2\%.
Their work also does not consider the directionality of the delta, and consequently they do not penalize situations where humans and metric disagree on which system is better.

\citet{mathur-etal-2020-tangled} found that even statistical significant deltas of up to three BLEU points do not reliably correspond to human judgement.
In a broad survey, \citet{marie-etal-2021-scientific} notes that various community ``rules of thumb'' about sufficient BLEU deltas might be the result of an evolved consensus that has no basis in scientific evidence. Similarly, 
\citet{kocmi-etal-2021-ship} demonstrated that among system pairs deemed statistically significant by humans and where \bleu disagree with humans, the median delta is 1.3 \bleu.
\citet{Marie2022} reinvestigated the WMT 2020 and 2021 results and showed that deltas lower than 2 \bleu needs to be tested for statistical significance.

Automated metrics in NLP and MT have been under scrutiny for a long time.
\citet{hovy-ravichandran-2003-holy} raised early doubts about BLEU.
\citet{callison-burch-etal-2006-evaluating} pointed to failure modes of BLEU and suggested it be used in more narrow situations.
\citet{post-2018-call} identified a problem with conflicting implementations of \bleu and offered a unified solution.
The broader field of computer science has been concerned with what is a meaningful acceptance threshold of a metric \citep{mori2018evaluating}.
The acceptance thresholds are usually established to trade off risks in types of errors \citep{shatnawi2010finding}.
\citet{kelley2012effect}, studying effect sizes in psychology, summarize that effect sizes should be scaled appropriately.
Alike, \citet{plonsky2014big} ask what effect size suffices and note its dependence on the variance and that all acceptance thresholds are arbitrary.

\section{Conclusion}

In this work, we investigated the interpretation of deltas from automatic machine translation metrics.
Although metrics have different ranges of scores, what ultimately matters to the practitioner is how score \emph{deltas} are grounded in human ability to perceive those differences, which we judge by pairwise system-level accuracy on a large collection of human judgments.
We empirically determined thresholds for popular metrics to align them on accuracy and provide a tool\footnotemark[0] that relates metrics to each other.
Finally, we showed the importance of using metric-delta accuracy over $p$-values: the former is stable across testset sizes.

We undertook some investigations into sub-factors of the data, showing that the results were robust to, for example, translation direction, and also that they generalized to different testsets.
These investigations were limited by the data size.
For future work, it would be useful to explore delta-accuracy for different subsets and combinations of features, presuming that enough data were available for the task.

\section*{Limitations}

While this work provides more informed guidelines on interpreting metric delta, they remain crude and do not fix the inadequacy of automated metrics.
In order to guarantee improvements, human evaluations need to be carried out.

We use humans as a gold standard, however, they are noisy and also unreliable especially for systems that are close in performance.

Almost all MT systems used in this meta-evaluation are not based on LLMs. Therefore, we may observe different behaviour of automatic metrics when evaluating LLM-based models.

Our estimated accuracy should not be used as the reason to reject a result, similarly as low significance p-value.

\section*{Ethics Statement}

The human annotators have been compensated considerably higher than the minimum wage standards in their respective countries. This commitment reflects our dedication to fair labor practices and the well-being of those contributing to our work.

\section*{Acknowledgements}
We would like to thank Arul Menezes, Roman Grundkiewicz, Martin N. Danka, Benjamin Marie and to the Microsoft Translator research team for their valuable feedback.

\bibliography{misc/anthology,misc/custom}

\appendix

\section{Metric Implementation Details}
\label{app:metrics_details}

There are many automatic metrics that has been developed. In our study, our selection of metrics has focus on either the most used metrics or the best performing ones. Here is the description, reason for their selection and details of implementations used.
For metric quality, please, refer to \Cref{sec:full_metric_evaluation} or \citet{kocmi-etal-2021-ship, freitag-etal-2023-results}.
However, we plan to extend the list of automatic metric even after paper publishing. For other metrics and models, see \url{https://github.com/kocmitom/MT-Thresholds/}.

We are aware that the list is heavily affected by COMET variant models. 
However, when investigating best performing metrics from \perscite{freitag-etal-2023-results}, we can see that most are either based on COMET framework, they are not publicly available, or build on propritetary LLMs.

Out of the lexical-based metrics, we select three of them, which are the most used.
However, we emphasize that these metrics should no longer be used for MT evaluation \parcite{freitag-etal-2022-results}. We use SacreBLEU \parcite{post-2018-call} in version 2.3.1 with default setting:
\begin{itemize}[left=0mm,topsep=0mm,noitemsep]
    \item \bleu{} \parcite{papineni-etal-2002-bleu}: the most popular and currently one of the worst performing MT metrics (we used a specific tokenizer for Japanese and Chinese as recommended)
    \item \chrf{} \parcite{popovic-2015-chrf}: second most popular lexical-based metric with better performance
    \item \spbleu{} \parcite{goyal-etal-2022-flores}: metric popular when evaluating on Flores testset
\end{itemize}

\medskip
\noindent
Two BLEURT models (commit \texttt{cebe7e6}): 
\begin{itemize}[left=0mm,topsep=0mm,noitemsep]
    \item \bleurtdef{} \parcite{sellam-etal-2020-bleurt}: the default model when using BLEURT framework called BLEURT-Tiny. It is important to note, that its performance is worse than BLEU (\Cref{sec:full_metric_evaluation}) and should not be used as authors suggest.
    \item \bleurt{} \parcite{pu-etal-2021-learning}: the best performing Bleurt model
\end{itemize}

\medskip
\noindent
We evaluate five Comet models (v\texttt{2.1.0}), the most popular metric framework aside \bleu{}:
\begin{itemize}[left=0mm,topsep=0mm,noitemsep]
    \item \comet[]{20}: most frequently used model and the default reference based model until the end of year 2023. The model name \texttt{wmt20-comet-da}.
    \item \comet[]{22}: currently the default reference-based model (\texttt{wmt22-comet-da}), outperforming \comet[]{20}.
    \item \comet[QE]{21}: we picked \texttt{wmt21-comet-qe-mqm} for its unusual behaviour of using very small delta while reaching high pairwise accuracy.
    \item \cometkiwi: \texttt{wmt22-cometkiwi-da} is the best quality estimation model.
    \item \xcometxxl: the best performing publicly available metric as evaluated by \citet{freitag-etal-2023-results}.
\end{itemize}

\section{ToShip23 Dataset Details}
\label{app:ToShip2}

For this work, we introduce and analyze an extended version of a non-public \toshiptwo dataset. The main changes of a dataset are almost twice as many system pairs as in \toshipone \parcite{kocmi-etal-2021-ship}; more than ten new domains and new parallel testsets; improved human evaluation protocol; and evaluating the latest state-of-the-art MT models.

The parallel testsets for evaluating MT models that we use in the extended part are mostly a collection of non-published human translated sentences. 
We focus on using testsets in authentic direction, from original source into human translated reference (avoiding reverse testsets whenever possible, \inparcite{toral-etal-2018-attaining}).
In contrast to \toshipone, which uses mainly two domains (news and speech), we extended the domains by more than ten.

We reduced the total number of languages in the \toshiptwo from 101 to 94. 
The removed languages are those which are not supported by either BERT \citep{devlin-etal-2019-bert} or XLM-RoBERTa \citep{conneau-etal-2020-unsupervised} -- language models used in the most popular metrics -- therefore, we could not include those languages in our analysis.

The MT systems being part of the dataset are coming from the same distribution as in \toshipone, but evaluating the most recent state-of-the-art models including a limited number of LLM based translations.
Lastly, we improved the human evaluation protocol, moved from source-based DA towards DA+SQM \citep{kocmi-etal-2022-findings}.

\section{Metrics Disagreement on Ranking}
\label{app:onebleu}

Automatic metrics often disagree on a ranking which system is better even for large enough deltas. We illustrate this phenomena in this section.

\begin{figure}[t]
    \vspace{-1mm}
    \centering
    \includegraphics[width=\linewidth]{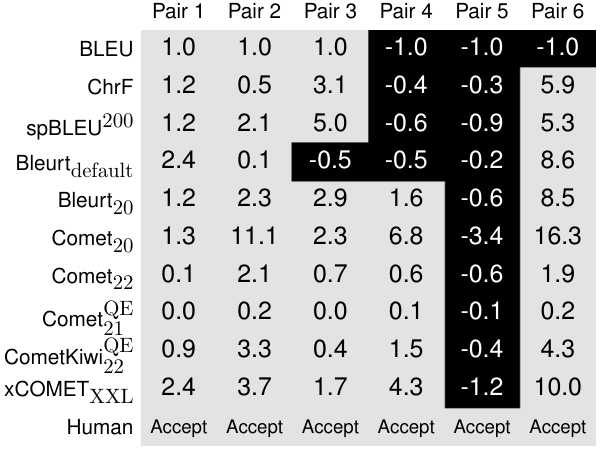}
    \caption{Subset of system pairs from WMT23 that have $\sim$1 BLEU delta. Each column is one system-pair. \textcolor{white}{\hlc[black]{\,Dark background\,}} represent metric disagreeing with humans on system ranking. This highlights that normalizing metrics towards BLEU range is not feasible.
    }
    \label{fig:onebleu_heatmap}
    \vspace{-2mm}
\end{figure}

We use the mostly unwritten (and long-debunked \citep{mathur-etal-2020-tangled}) operating assumption that +1--2 BLEU points denotes a significant finding as an anchor point to illustrate the range of metric deltas on a subset of systems in \Cref{fig:onebleu_heatmap}.
This figure reports metric deltas for six randomly-selected system pairs from WMT23 data, whose delta was roughly 1~\bleu{}.

As we can see in \Cref{fig:onebleu_heatmap}, while for first two system pairs, all metrics and humans agree on the system ranking, it is not the case for later four system pairs. For example, even \comet{20} score of 3.4 (fifth system pair) may result in disagreement with humans.

\section{Number of System Pairs in a Bin}
\label{app:bucket_size}

\begin{figure}[htbp]
  \centering
  \includegraphics[width=\linewidth]{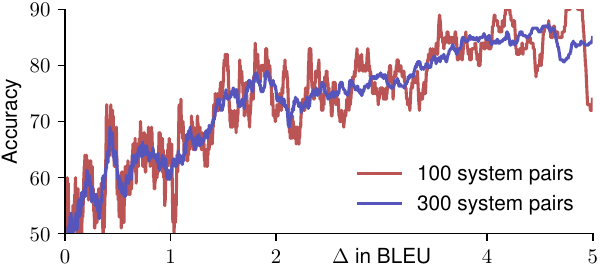}
  \includegraphics[width=\linewidth]{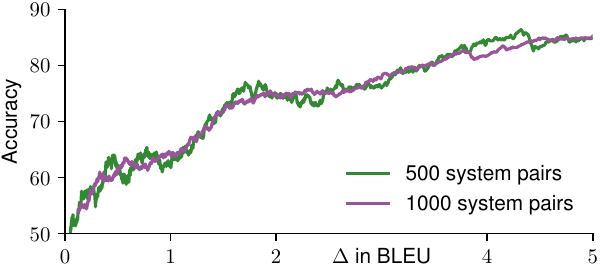}
  \caption{
  Comparison of pairwise accuracy for \bleu{} on \toshiptwo dataset when we change how many system pairs are in evaluation for each individual delta.
  }
  \label{fig:bucket_sizes}
\end{figure}

In our work, we fixed the number of systems in a bin for given metric delta to 300 system pairs.
We now show how this decision affected our evaluation.
To this end, we show various bin sizes in \Cref{fig:bucket_sizes}.
The bin width works as a smoothing parameter.
With bin size of 100 system pairs, the curve fluctuates, especially as one system pair transfer into 1\% change on the accuracy scale. 

We set the parameter to 300 system pairs because that is already a smoother curve, while not too wide so that the epsilon around the investigated delta is also not too high. 
However, this parameter should be re-investigated in the future works.

\begin{figure}[htbp]
  \centering
  \includegraphics[width=\linewidth]{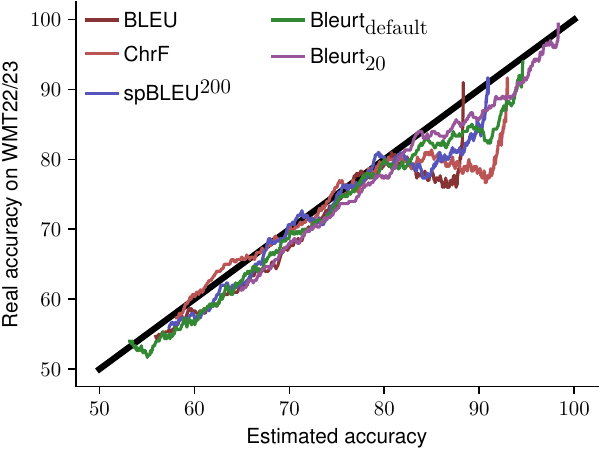}
  \includegraphics[width=\linewidth]{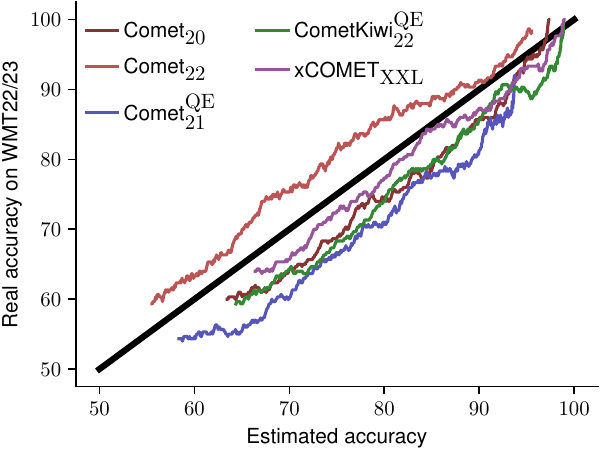}
  \caption{
  Testing the validity of thresholds devised on \toshiptwo with WMT datasets. In a scenario without noisy data, we would expect the real accuracies to match the estimated accuracies (the black line).
  This figure provides more detail on \Cref{fig:test_on_wmt}.
  }
  \label{fig:test_on_wmt_big}
\end{figure}

\begin{figure*}[htbp]
  \centering
  \includegraphics[width=0.24\linewidth]{generated/empirical_accuracies/,BLEU_sigmoid.pdf} 
  \includegraphics[width=0.24\linewidth]{generated/empirical_accuracies/,ChrF_sigmoid.pdf} 
  \includegraphics[width=0.24\linewidth]{generated/empirical_accuracies/,BLEURT20_sigmoid.pdf} 
  \includegraphics[width=0.24\linewidth]{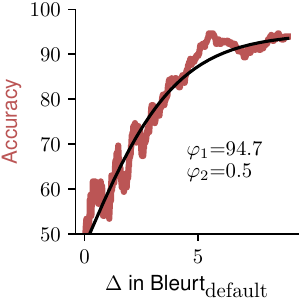} 
  \raisebox{-1mm}{\includegraphics[width=0.24\linewidth]{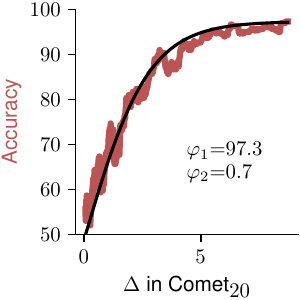}}
  \includegraphics[width=0.24\linewidth]{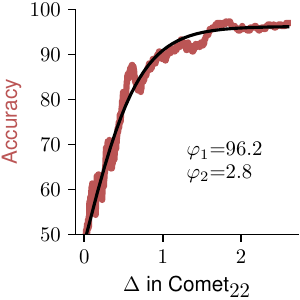} 
  \raisebox{-1mm}{\includegraphics[width=0.24\linewidth]{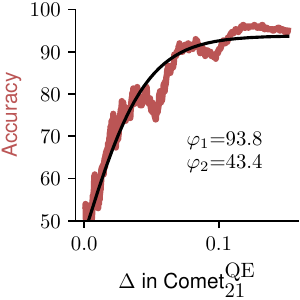}}
  \raisebox{-1mm}{\includegraphics[width=0.24\linewidth]{generated/empirical_accuracies/,COMETKIWI22-QE-src_sigmoid.pdf}}
  \includegraphics[width=0.24\linewidth]{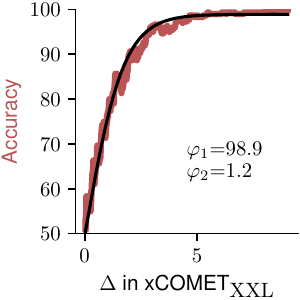}
  \includegraphics[width=0.24\linewidth]{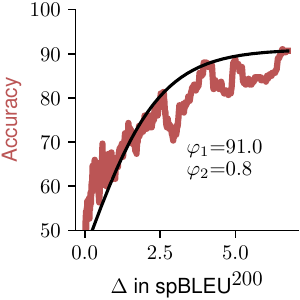} 
  \caption{
  Empirical pairwise \textcolor{RedA}{accuracies} for all metrics with a fitted sigmoid curves on \toshiptwo dataset. This figure extends \cref{fig:expected-accuracy-with_sigmoid}.
  }
  \label{fig:expected-accuracy-with_sigmoid_big}
\end{figure*}

\begin{figure*}[t]
  \centering
  \includegraphics[width=0.47\linewidth]{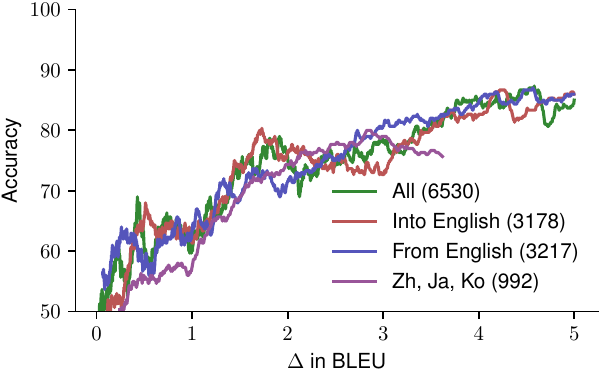}
    \includegraphics[width=0.47\linewidth]{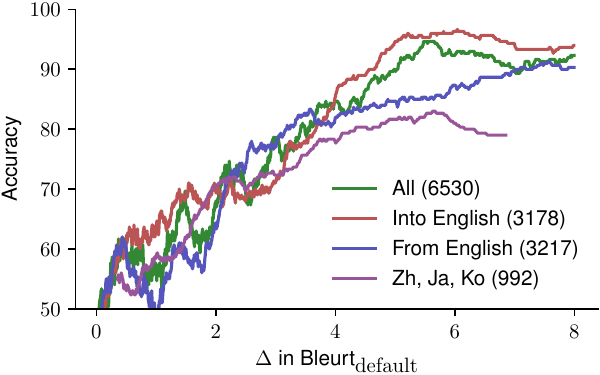}
    \includegraphics[width=0.47\linewidth]{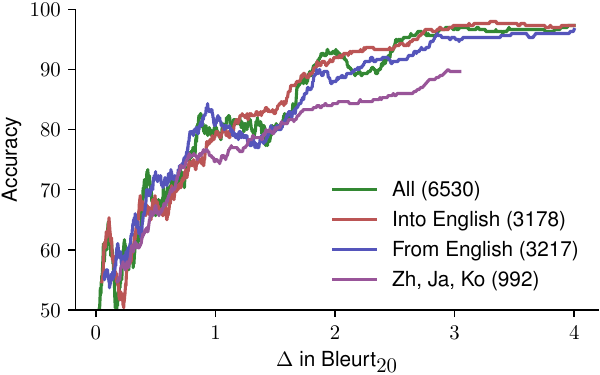}
      \includegraphics[width=0.47\linewidth]{generated/XE_EX/,ChrF.pdf}

    \includegraphics[width=0.47\linewidth]{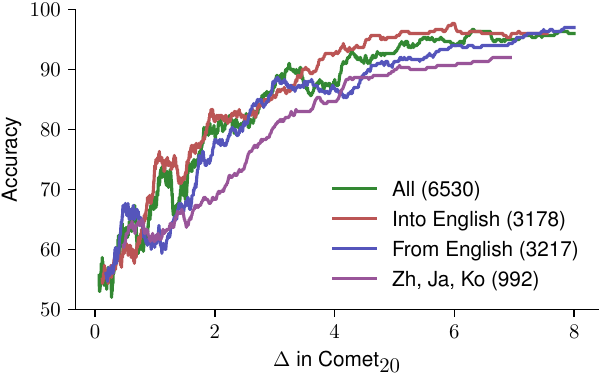}
    \includegraphics[width=0.47\linewidth]{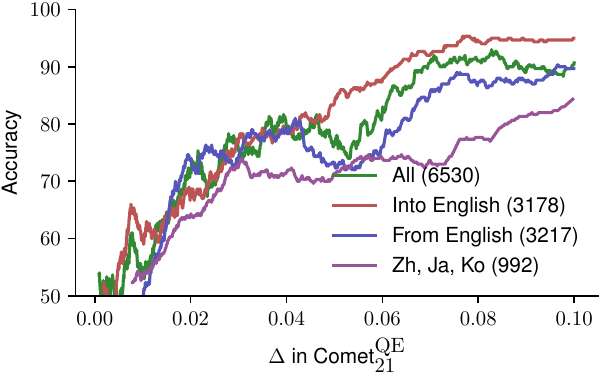}
    \includegraphics[width=0.47\linewidth]{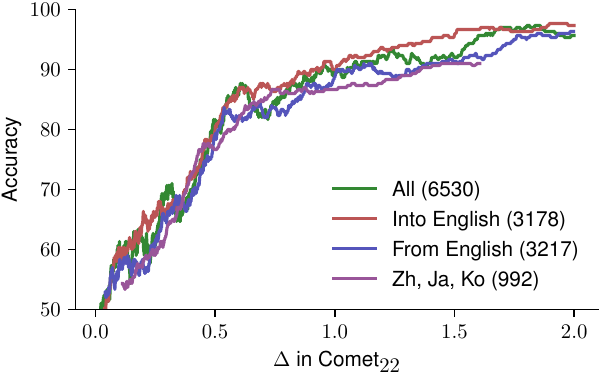}
    \includegraphics[width=0.47\linewidth]{generated/XE_EX/,COMETKIWI22-QE-src.pdf}
    \includegraphics[width=0.47\linewidth]{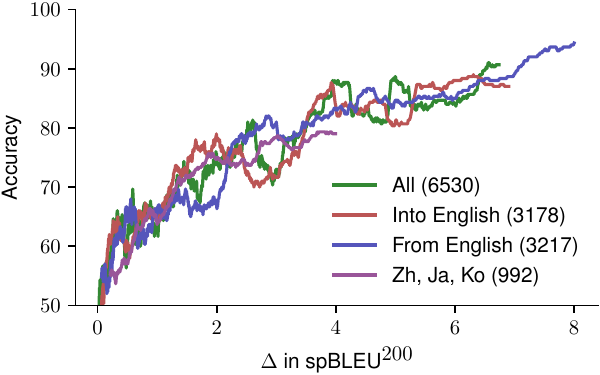}
    \includegraphics[width=0.47\linewidth]{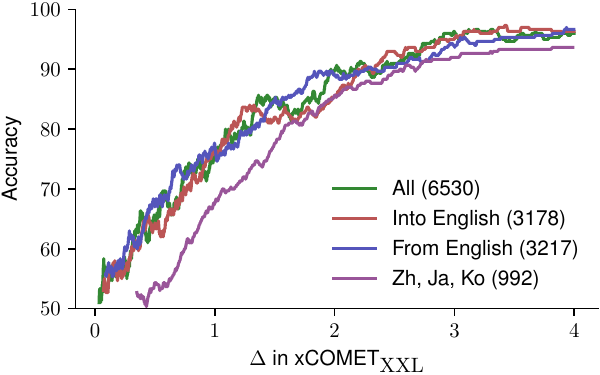}
  \caption{
  Comparison of pairwise accuracy on \toshiptwo dataset when comparing into English, out-of-English, and Chinese, Japanese, Korean language pairs separately. The count shows total number of system-pairs in the evaluation.
  This figure extends \Cref{fig:XE_EX_comparison}. 
  }
  \label{fig:XE_EX_comparison_big}
\end{figure*}

\begin{figure*}[t]
  \centering
  \includegraphics[width=0.47\linewidth]{generated/unrelated_systems/,BLEU.pdf}
  \includegraphics[width=0.47\linewidth]{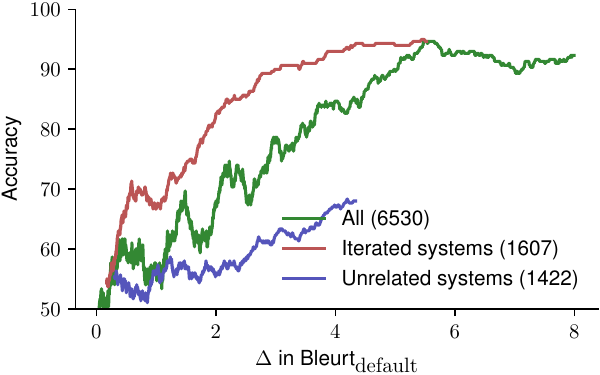}
  \includegraphics[width=0.47\linewidth]{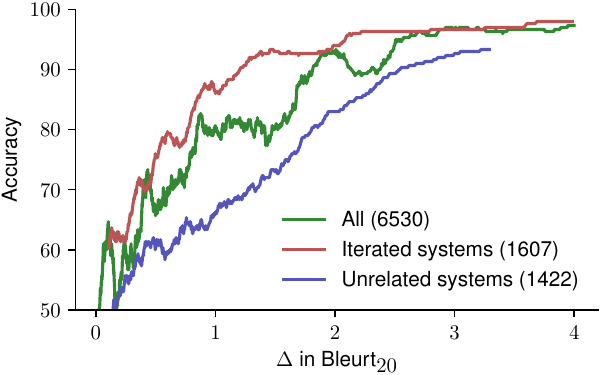}
  \includegraphics[width=0.47\linewidth]{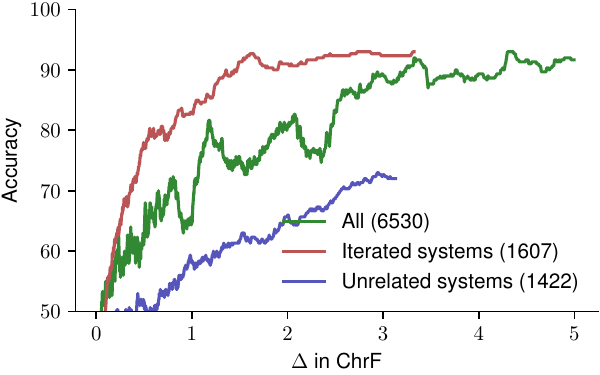}
  \includegraphics[width=0.47\linewidth]{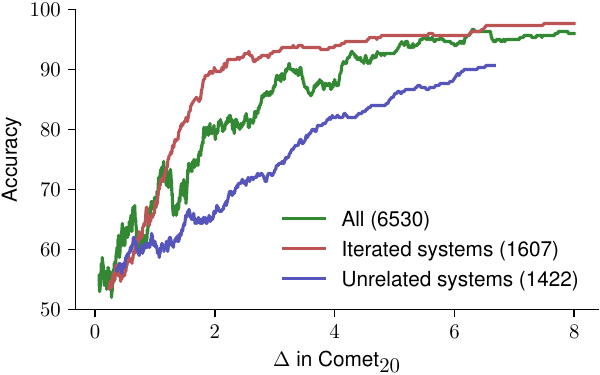}
  \includegraphics[width=0.47\linewidth]{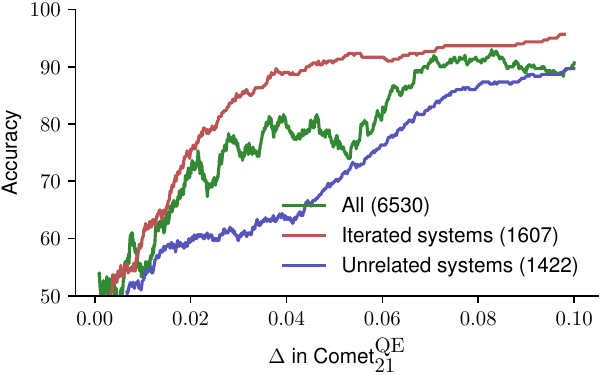}
  \includegraphics[width=0.47\linewidth]{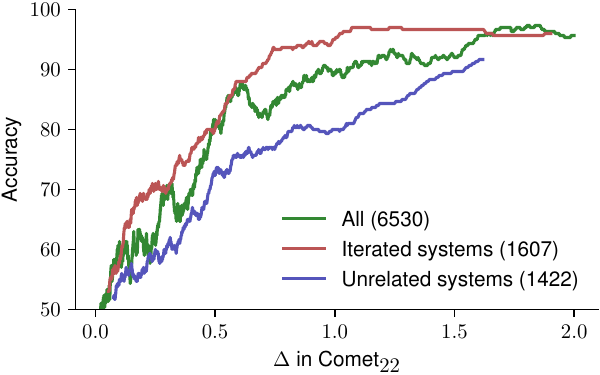}
  \includegraphics[width=0.47\linewidth]{generated/unrelated_systems/,COMETKIWI22-QE-src.pdf}
  \includegraphics[width=0.47\linewidth]{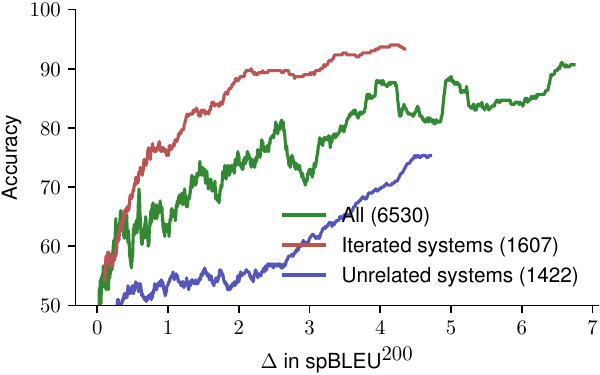}
  \includegraphics[width=0.47\linewidth]{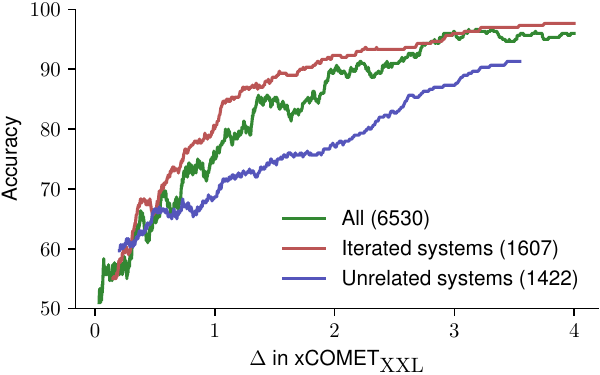}
  \caption{
  Comparison between iterated and unrelated systems on \toshiptwo.
  This figure extends \Cref{fig:unrelated_systems}.
  }
  \label{fig:unrelated_systems_big}
\end{figure*}

\end{document}

%% file: generated/dataset_overview.tex
\begin{tabular}{lrrrrr}
\toprule
\bf Dataset & \bf Segments & \bf Systems & \bf Sys. pairs & \bf Langs. & \bf Domains\\
\midrule
WMT22 & 221k & 108 & 543 & 8 & 4\\
WMT23 & 223k & 129 & 871 & 7 & 4\\
ToShip21 & 2300k & 4380 & 3344 & 101 & 2\\
ToShip23 & 3016k & 6752 & 6530 & 94 & >10\\
\bottomrule
\end{tabular}

%% file: generated/main_thresholds.tex
\begin{tabular}{lllllllllll}
\toprule
\bf \makecell[l]{Estimated \\ Accuracy} & \bf \makecell[c]{{\small Coin toss}\\ 50\%} & \bf 55\% & \bf 60\% & \bf 65\% & \bf 70\% & \bf 75\% & \bf 80\% & \bf 85\% & \bf 90\% & \bf 95\% \\
\midrule
BLEU                                &                                        0.27 &     0.52 &     0.78 &     1.06 &     1.39 &     1.79 &     2.34 &     3.35 &        - &        - \\
ChrF                                &                                        0.14 &     0.33 &     0.54 &     0.76 &     1.00 &     1.28 &     1.63 &     2.12 &     3.05 &        - \\
spBLEU$^\textrm{200}$               &                                        0.25 &     0.52 &     0.82 &     1.13 &     1.49 &     1.91 &     2.46 &     3.28 &     5.57 &        - \\
Bleurt$_\textrm{default}$           &                                        0.23 &     0.66 &     1.11 &     1.59 &     2.11 &     2.71 &     3.43 &     4.39 &     5.98 &        - \\
Bleurt$_\textrm{20}$                &                                        0.02 &     0.17 &     0.33 &     0.49 &     0.66 &     0.85 &     1.07 &     1.35 &     1.73 &     2.44 \\
Comet$_\textrm{20}$                 &                                        0.08 &     0.36 &     0.65 &     0.96 &     1.29 &     1.67 &     2.10 &     2.66 &     3.45 &     5.10 \\
Comet$_\textrm{22}$                 &                                        0.03 &     0.10 &     0.18 &     0.26 &     0.35 &     0.45 &     0.56 &     0.71 &     0.94 &     1.53 \\
Comet$_\textrm{21}^\textrm{QE}$     &                                       0.003 &    0.008 &    0.013 &    0.019 &    0.025 &    0.032 &    0.041 &    0.052 &    0.073 &        - \\
CometKiwi$_\textrm{22}^\textrm{QE}$ &                                        0.01 &     0.08 &     0.16 &     0.24 &     0.33 &     0.42 &     0.53 &     0.67 &     0.85 &     1.18 \\
xCOMET$_\textrm{XXL}$               &                                        0.02 &     0.19 &     0.37 &     0.56 &     0.76 &     0.98 &     1.24 &     1.55 &     1.99 &     2.74 \\
\bottomrule
\end{tabular}

%% file: generated/accuracy_results.tex
\begin{tabular}{lcccc}
\toprule
{} & \bf ToShip23 & \bf 22-23 & \bf 19-21 & \bf WMT23 \\
\midrule
system pairs (N)                    &         6530 &      1843 &      4687 &       249 \\
CometKiwi$_\textrm{22}^\textrm{QE}$ &         81.5 &      74.5 &      84.3 &      90.0 \\
xCOMET$_\textrm{XXL}$               &         81.4 &      75.3 &      83.9 &      92.8 \\
Comet$_\textrm{20}$                 &         80.1 &      73.2 &      82.9 &      86.3 \\
Bleurt$_\textrm{20}$                &         78.6 &      69.8 &      82.1 &      89.2 \\
Comet$_\textrm{22}$                 &         78.6 &      71.1 &      81.5 &      84.7 \\
Comet$_\textrm{21}^\textrm{QE}$     &         76.8 &      71.2 &      79.0 &      69.5 \\
ChrF                                &         71.9 &      61.4 &      76.0 &      79.5 \\
spBLEU$^\textrm{200}$               &         71.6 &      61.0 &      75.7 &      81.9 \\
BLEU                                &         70.3 &      61.3 &      73.9 &      81.5 \\
Bleurt$_\textrm{default}$           &         69.9 &      61.0 &      73.4 &      85.1 \\
\bottomrule
\end{tabular}